%% file: main.tex
\documentclass[11pt]{article}
\usepackage{times}

\usepackage{hyperref}

\usepackage{xcolor}
\definecolor{defaultcolor}{gray}{0.9}
\newlength\savewidth

\usepackage[symbol]{footmisc}
\usepackage[noblocks]{authblk}

\usepackage{booktabs}
\usepackage{graphicx}
\usepackage{multirow}
\usepackage{color, colortbl}
\usepackage{caption}
\usepackage{subcaption}
\usepackage{wrapfig}
\usepackage{makecell}
\usepackage{amsfonts}
\usepackage{amsmath}
\usepackage{booktabs}
\usepackage{xspace}
\usepackage{array}
\usepackage{siunitx}
\usepackage{enumitem}
\usepackage[most]{tcolorbox}
\usepackage[square, numbers, sort&compress]{natbib}
\usepackage[margin=1in]{geometry}
\usepackage[normalem]{ulem}
\usepackage{fancyhdr}
\usepackage{soul}

\newcommand{\eat}[1]{\ignorespaces}

\newcolumntype{F}[1]{%
    >{\raggedright\arraybackslash\hspace{0pt}}p{#1}}%
\newcolumntype{T}[1]{%
    >{\centering\arraybackslash\hspace{0pt}}p{#1}}%

\usepackage[margin=1in]{geometry}
\usepackage[symbol]{footmisc}

\def\ourmethod{\textit{BiomedParse}\xspace}
\def\ourdata{\textit{BiomedParseData}\xspace}

\title{\ourmethod: a biomedical foundation model for image parsing of everything everywhere all at once}

\author{\textbf{Theodore Zhao$^{1*\dagger\P}$, Yu Gu$^{1*}$, 
Jianwei Yang$^{1}$, Naoto Usuyama$^{1}$, Ho Hin Lee$^{1}$, Tristan Naumann$^{1}$, Jianfeng Gao$^{1}$,  
Angela Crabtree$^{2}$, Jacob Abel$^{2}$, Christine Moung-Wen$^{2}$, Brian Piening$^{2}$, Carlo Bifulco$^{2}$, Mu Wei$^{1, \ddag}$, Hoifung Poon$^{1, \ddag, \S}$, Sheng Wang$^{3,\ddag}$}\\

$^{1}$Microsoft Research, Redmond, WA, USA\vspace{-1mm}\\
$^{2}$Providence Genomics, Portland, OR, USA\vspace{-1mm}\\    
$^{3}$Paul G. Allen School of Computer Science and Engineering, University of Washington, Seattle, WA, USA\vspace{-1mm}\\   
}
\footnotetext{$*$ Joint first authors}
\footnotetext{$\dagger$ Project lead}
\footnotetext{$\P$ Main technical contribution}
\footnotetext{$\ddag$ Corresponding authors: muhsin.wei@microsoft.com, swang@cs.washington.edu, hoifung@microsoft.com}
\footnotetext{$\S$ Lead contact}

\date{}

\begin{document}
\maketitle

\begin{center}
    \url{https://aka.ms/biomedparse-project}
\end{center}

\input{nbt_sec/abstract}

\newpage
\input{nbt_sec/introduction}

\input{nbt_sec/results}

\input{nbt_sec/discussion}
\input{nbt_sec/methods}

\input{data_code_arxiv}

\newpage
\bibliography{references}
\bibliographystyle{naturemag}

\input{nbt_sec/figures}

\input{nbt_sec/supplementary}

\end{document}

%% file: nbt_sec/abstract.tex
\begin{abstract}

\noindent Biomedical image analysis is fundamental for biomedical discovery in cell biology, pathology, radiology, and many other biomedical domains. Holistic image analysis comprises interdependent subtasks such as segmentation, detection, and recognition of relevant objects. Traditionally, these tasks are tackled separately. For example, there have been a lot of works focusing on segmentation alone, completely ignoring key semantic information in downstream tasks of detection and recognition. In contrast, image parsing is a unifying framework that jointly pursues these tasks by leveraging their interdependencies such as the semantic label of a segmented object. Here, we propose \ourmethod, a biomedical foundation model for imaging parsing that can jointly conduct segmentation, detection, and recognition for 82 object types across 9 imaging modalities. Through joint learning, we can improve accuracy for individual tasks and enable novel applications such as segmenting all relevant objects in an image through a text prompt, rather than requiring users to laboriously specify the bounding box for each object. Interestingly, we can train \ourmethod using no more than standard segmentation datasets. The key is to leverage readily available natural-language labels or descriptions accompanying those datasets and use GPT-4 to harmonize the noisy, unstructured text information with established biomedical object ontologies. We created a large dataset comprising over six million triples of image, segmentation mask, and textual description. On image segmentation, we showed that \ourmethod is broadly applicable, outperforming state-of-the-art methods on 102,855 test image-mask-label triples across 9 imaging modalities (\textit{everything}). \ourmethod is also able to identify invalid user inputs describing objects that do not exist in the image. On object detection, which aims to locate a specific object of interest, \ourmethod again attained state-of-the-art performance, especially on objects with irregular shapes (\textit{everywhere}). On object recognition, which aims to identify all objects in a given image along with their semantic types, we showed that \ourmethod can simultaneously segment and label all biomedical objects in an image (\textit{all at once}). In summary, \ourmethod is an all-in-one tool for biomedical image analysis by jointly solving segmentation, detection, and recognition. It is broadly applicable to all major biomedical image modalities, paving the path for efficient and accurate image-based biomedical discovery.

\end{abstract}

%% file: nbt_sec/introduction.tex
\section*{Introduction}
Biomedical image analysis is critical to biomedical discovery because imaging is one of the most important tools for studying physiology, anatomy, and function at multiple scales from the organelle level to the organ level \cite{royer2023future, li2023challenges}. Holistic image analysis comprises multiple subtasks, such as segmentation, detection, and recognition of biomedical objects. Segmentation aims to divide an image into segments representing different objects, often requiring the aid of a user-provided bounding box for each object of interest \cite{wang2022medical, salpea2022medical}. Detection aims to identify the location of an object of interest in the image \cite{ribli2018detecting}, whereas recognition aims to identify all objects within an image \cite{ma2023cellcano}. 
Standard image analysis methods typically approach these tasks separately, using specialized tools for individual tasks \cite{jiang2023review}. Despite their encouraging performance, such a disjoint approach misses significant opportunities for joint learning and reasoning across these interdependent tasks. 

For example, a lot of prior image analysis works focus on segmentation alone, thus ignoring key semantic information from downstream tasks of object detection and recognition. This results in sub-optimal segmentation while imposing substantial burden on users, as many state-of-the-art segmentation tools require users to provide a tight bounding box indicating the location of an object of interest \cite{kirillov2023segment, ma2024segment}. The bounding-box requirement leads to three limitations. First, users have to manually draw bounding boxes in the image, which requires domain expertise to identify the locations and shapes of the target objects. Second, bounding boxes, which are often rectangular, fall short of accurately representing objects with irregular or complex shapes. Third, bounding box-based approaches are not scalable for images containing a large number of objects, such as segmenting cells in a whole-slide pathology image, since users need to provide a bounding box for each object. 

In this paper, we propose to approach biomedical image analysis as {\em image parsing}, a unifying framework for joint learning and reasoning across segmentation, detection, and recognition ~\cite{tu2005image, tighe2013superparsing, zhou2015medical}. Specifically, we have developed \ourmethod, a biomedical foundation model for image parsing that is capable of carrying out all three tasks by leveraging their interdependencies, thus addressing key limitations in traditional methods. In particular, joint learning of object detection and recognition eliminates the need for user-specified bounding boxes, as segmentation can be done using semantic labels from text prompt alone. 

The major bottleneck for pretraining \ourmethod is data. While biomedical segmentation datasets abound ~\cite{gamper2020pannuke, ji2022amos, ACDCData}, there are relatively few prior works on object detection and recognition in biomedicine, let alone datasets covering all three tasks. 
To address this problem, we propose a novel approach for pretraining \ourmethod using no more than standard segmentation datasets. The key insight is to leverage readily available natural-language labels or descriptions accompanying those datasets and use GPT-4 to harmonize these noisy, unstructured texts with established biomedical object ontologies. This enables us to construct \ourdata, a biomedical image parsing dataset comprising 3.4 million triples of image, segmentation mask, and semantic label of the biomedical object and 6.8 million image-mask-description triples, from over 1 million images. The semantic labels encompass 82 major biomedical object types across 9 imaging modalities.

Unlike segmentation methods that focus on identifying salient segment boundary within a bounding box, \ourmethod learns to model typical shape of each object class, thus mimicking how humans perceive objects in an image. \ourmethod can segment images using text prompts alone (e.g. ``inflammatory cells in breast pathology''), without requiring any user-specified localization such as bounding boxes. Consequently, \ourmethod can better recognize and segment objects of irregular and complex shapes, which are very challenging for traditional methods using rectangular bounding boxes. Moreover, \ourmethod can recognize all objects in an image, without requiring any user input prompt.


We conduct a large-scale study to evaluate \ourmethod on 102,855 held-out image-mask-label triples across 9 modalities for segmentation, detection, and recognition. 
On segmentation, \ourmethod established new state-of-the-art results, outperforming prior best methods such as MedSAM \cite{ma2024segment} and SAM \cite{kirillov2023segment}. 
Moreover, using text prompts alone, \ourmethod is much more scalable than these prior methods that require orders of magnitude more user operations in specifying object-specific bounding boxes to perform competitively. 
We also demonstrated that \ourmethod can accurately detect invalid text prompts describing non-existent objects in the image. 
On detection, we show that \ourmethod learns accurate modeling of all object classes, including those with irregular shapes. This results in even larger improvement in image analysis accuracy for such objects, attaining a 0.857 Dice score that is 39.6\% higher than the best-competing method. On recognition, we show how \ourmethod can accurately segment and label all objects without any user-specified text prompt. Collectively, we introduce a biomedical foundation model for image parsing, achieving superior performance on segmentation, detection, and recognition, paving the path for large-scale image-based biomedical discovery.

%% file: nbt_sec/results.tex
\section*{Results}
\subsection*{Overview of \ourmethod and \ourdata}

To develop a model that can jointly conduct segmentation, detection, and recognition, we need a supervision dataset that covers all three tasks. To our best knowledge, no such a dataset exists. 
To this end, we created the first such dataset \ourdata by combining 45 biomedical image segmentation datasets and using GPT-4 to generate the canonical semantic label for each segmented object. 

The key insight is that existing segmentation datasets often contain valuable semantic information about the segmented objects. However, such information typically resides in noisy and inconsistent natural-language text descriptions that do not conform to standard biomedical ontologies. 
To address this challenge, we use GPT-4 to create a unifying biomedical object ontology for image analysis and harmonize natural-language descriptions with this ontology (see \textbf{Methods}).
This ontology encompasses three main categories (histology, organ, abnormality), 15 meta-object types, and 82 specific object types (\textbf{Fig. \ref{fig1:overview}a}).

The resulting \ourdata contains 3.4 million distinct image-mask-label triples, spanning 9 imaging modalities and 25 anatomic sites (\textbf{Fig. \ref{fig1:overview}b, \textbf{Supplementary Figure} \ref{fig:data-modality-chord}}), representing a large-scale and diverse dataset for semantic-based biomedical image analysis. 

To make \ourmethod better equipped in handling diverse text prompts not covered by the canonical semantic labels, we also use GPT-4 to synthesize synonymous text descriptions for each semantic label and sample from them during training. This yielded a total of 6.8 million image-mask-description triples (see \textbf{Methods}, \textbf{Supplementary Figure} \ref{fig:prompt} and \ref{fig:example}). 

While our method does not use bounding boxes, prior state-of-the-art methods such as MedSAM and SAM generally require pre-specified bounding boxes. We consider two scenarios to provide the bounding boxes: oracle bounding box (the minimum rectangular bounding box covering a segmented object) and bounding box created by Grounding DINO \cite{liu2023grounding}, a state-of-the-art object detection method that can generate bounding boxes from text prompt of an object label. (Grounding DINO does not perform segmentation.)

\ourmethod adopts a modular design under the SEEM architecture \cite{zou2024segment}, comprising an image encoder (for encoding the input image), a text encoder (for encoding the text prompt), a mask decoder (for outputing segmentation mask), and a meta-object classifier (for joint training of image encoder with object semantics). See \textbf{Fig. \ref{fig1:overview}c}. The image and text encoders were initialized using state-of-the-art Focal~\cite{yang2022focal} and PubMedBERT ~\cite{gu2021domain}, respectively. 

Before evaluating image analysis results, we first examine the quality of embeddings derived from \ourmethod.
Specifically, we compare the text embeddings from \ourmethod with those from PubMedBERT. 
We found that embeddings from \ourmethod can better distinguish fine-grained cell types, with a Silhouette score of 0.89 that is much higher than using the embeddings from PubMedBERT (\textbf{Fig. \ref{fig1:overview}d}, \textbf{Supplementary Figure} \ref{fig:ablation}). We also compare the image embeddings from \ourmethod with those from Focal. We observed that embeddings from \ourmethod are more predictive of tumor malignancy on a pathology dataset \cite{glas} (\textbf{Fig. \ref{fig1:overview}e}). The superior performance of the text and image embeddings from \ourmethod necessitates the training of \ourmethod using \ourdata, raising our confidence that \ourmethod can be an effective approach for biomedical image analysis.

\subsection*{Accurate and scalable segmentation across nine modalities}

We first evaluated \ourmethod on biomedical image segmentation using the held-out set comprising 102,855 test instances (image-mask-label triples) across 9 imaging modalities (\textbf{Fig. \ref{fig2:quant_eval}a}, \textbf{Supplementary Figure} \ref{fig:ablation}). We observed that \ourmethod achieved the best Dice score, even against the best-competing method MedSAM with oracle bounding box as input (paired t-test \textit{p}-value $< 10^{-4}$). 
In the more realistic setting when MedSAM or SAM is supplied with bounding boxes generated by Grounding DINO, the superiority of \ourmethod is even more prominent in end-to-end biomedical object detection and segmentation, especially in more challenging modalities such as pathology and CT where irregular-shaped objects abound. 
By training on domain-specific datasets, both \ourmethod and MedSAM outperform general-domain methods such as SAM. We showed examples comparing \ourmethod segmentation and the ground truth across multiple imaging modalities, demonstrating the generalizability of \ourmethod (\textbf{Fig. \ref{fig2:quant_eval}b}). We further compared \ourmethod on a benchmark created by MedSAM \cite{ma2024segment} encompassing 50 tasks and again observed the best performance by \ourmethod, even against MedSAM with oracle bounding box (paired t-test \textit{p}-value $< 10^{-2}$), further demonstrating the superiority of \ourmethod (\textbf{Supplementary Figure} \ref{fig:overlap_eval_mod}).

In addition to being more accurate, \ourmethod is more scalable compared to bounding box-based approaches, which stems from the generalizability of text prompts across images of the same modality or anatomical site, thus eliminating the need for laborious user operations in providing a tight bounding box for each object. 
To demonstrate this, we compared \ourmethod and prior state-of-the-art methods MedSAM and SAM on a cell segmentation dataset with 42 colon pathology images (\textbf{Fig. \ref{fig2:quant_eval}c}). Using a single text prompt ``glandular structure in colon pathology image'', \ourmethod achieves a 0.942 median Dice score, whereas neither SAM nor MedSAM achieves a median Dice score higher than 0.75 without tight bounding boxes as input. 
In fact, to achieve competitive results comparable to \ourmethod with a single text prompt, MedSAM requires the users to supply a tight bounding box for each of the 430 cells in these images (\textbf{Fig. \ref{fig2:quant_eval}c}). 
In general, our results reveal that the bounding box-based approach is much less accurate on irregular-shaped objects, such as tumors and abnormal cells (\textbf{Fig. \ref{fig2:quant_eval}d,e}). In contrast, \ourmethod still attained highly accurate segmentation for such objects. The scalability and accuracy of \ourmethod bode well for its utility in real-world applications.

\ourmethod can also detect invalid text prompts (e.g., the request to identify a brain tissue in a chest X-Ray image), by calculating a \textit{p}-value using Kolmogorov–Smirnov (K-S) test (see \textbf{Methods}). From preliminary experiments, we found that invalid text prompts have an average K-S test $p$-value smaller than $10^{-3}$ while the valid ones have an average K-S test $p$-value above 0.1 (\textbf{Fig. \ref{fig2:quant_eval}f}). Using 0.01 as the p-value cutoff, \ourmethod can achieve an estimated performance of 0.93 precision and 1.00 recall on detecting invalid input (\textbf{Fig. \ref{fig2:quant_eval}g}). \ourmethod substantially outperformed Grounding DINO on invalid input detection (AUROC 0.99 vs 0.61). See \textbf{Fig. \ref{fig2:quant_eval}h,i}). This enables \ourmethod to perform recognition by enumerating candidate object types in the ontology, skipping invalid text prompts and generating segmentation masks for valid object labels.

\subsection*{Accurate detection of irregular-shaped objects}

Next, we evaluated the performance of \ourmethod on object detection, where the model is asked to identify an object of interest in the image. \ourmethod can resolve natural-language variations and accept text prompts that do not exactly match any semantic label in the ontology. 
In the previous section, we already show that \ourmethod outperformed bounding-box-based methods in general. Additionally, since \ourmethod learns semantic representation for individual object types, We hypothesize that its superiority over prior methods will be even more pronounced in detecting irregular-shaped objects. To verify this, we show the aggregate attention map of each object type learned by \ourmethod on test images unseen during training and observed that they faithfully reflect object shapes, including many irregular-shaped objects (\textbf{Fig. \ref{fig3:irreg_eval}a}). Next, we define three metrics to assess the regularity of an object, including \textit{convex ratio} (i.e., the ratio of the object size to the tightest convex size), \textit{box ratio} (i.e., the ratio of the object size to the tightest rectangle size), and \textit{rotational inertia} (i.e., the difficulty in changing the rotational velocity) (see \textbf{Methods}). We found that the improvements of \ourmethod over SAM and MedSAM are strongly correlated with these metrics (average correlation 0.829), indicating that our method has a larger improvement on irregular-shaped objects (\textbf{Fig. \ref{fig3:irreg_eval}b-d, Supplementary Figure \ref{fig:irregular_medsam}}). \textbf{Fig. \ref{fig3:irreg_eval}e} illustrates a few examples comparing \ourmethod and MedSAM on detecting irregular-shaped objects. Furthermore, we show that \ourdata has higher average object irregularity than the datasets used by MedSAM (\textbf{Fig. \ref{fig3:irreg_eval}f,g}, \textbf{Supplementary Figure} \ref{fig:IRI_violin}), and the improvement of \ourmethod is also larger on \ourdata (\textbf{Fig. \ref{fig3:irreg_eval}h}), highlighting the benefit from joint learning of object semantics in detecting the more challenging irregular-shaped objects. 

\subsection*{Object recognition using the segmentation ontology}

In our final analysis, we explore \ourmethod's capacity for object recognition, which aims to simultaneously segment and label every object within an image. Provided with an image, along with its modality and anatomical site, \ourmethod iteratively performs detection and segmentation for all candidate object types within the ontology of that modality and anatomical site, and the segmented masks are aggregated to ensure spatial cohesion among adjacent pixels (see \textbf{Methods}). This enables \ourmethod to accurately conduct object recognition, as evidenced in \textbf{Fig. \ref{fig4:pano_eval}a}, where objects are accurately identified and segmented with an average Dice score of 0.94.

Grounding DINO \cite{liu2023grounding} is the state-of-the-art general-domain object recognition system but it does not perform segmentation, which makes Grounding DINO and \ourmethod not directly comparable. We circumvent this by casting the object recognition task as a binary classification problem: given an input image and a candidate object type, the model determines whether the image contains at least one object of the given type. In this classification formulation, we observed that \ourmethod substantially outperformed Grounding DINO with a 25.0\%, 87.9\%, 74.5\% improvement on precision, recall, and F-1, respectively (\textbf{Fig. \ref{fig4:pano_eval}b-d}).  The improvement over Grounding DINO is even larger when more objects are present in the image (\textbf{Fig. \ref{fig4:pano_eval}e}).


Next, we evaluated the performance of \ourmethod on end-to-end object recognition using weighted average Dice score. 
Compared with MedSAM and SAM using Grounding DINO for recognition and bounding box generation, \ourmethod outperformed them by a large margin (\textbf{Fig. \ref{fig4:pano_eval}f, Supplementary Figure \ref{fig:panoptic_scatter}}). Similar to our observation on object identification, the improvement over comparison approaches is even larger when more objects are present in the image (\textbf{Fig. \ref{fig4:pano_eval}g}). These results indicate \ourmethod's ability to identify all objects in an image, offering an effective tool for holistic image analysis.

Finally, we evaluated \ourmethod on real-world data from the Providence Health System (\textbf{Fig. \ref{fig5:prov_ex}}). We performed object recognition here by asking \ourmethod to identify and segment all relevant cells in the pathology slides. We found that the annotations by \ourmethod correctly identified regions of immune cells and cancer cells, attaining high consistency with the pathologist annotations. While pathologists tend to focus on a specific region of cell type and provide coarse-grained annotations, \ourmethod can precisely label all relevant cells as specified in the ontology, indicating the potential for \ourmethod to help alleviate clinician burdens in real-world clinical applications.

%% file: nbt_sec/discussion.tex
\section*{Discussion}
We have presented \ourmethod, a biomedical foundation model for image analysis based on image parsing, and a large-scale image parsing dataset \ourdata containing 3.4 million image-mask-label triples and 6.8 million image-mask-description triples. In contrast to existing biomedical foundational models that require users to provide a tight bounding box for each object to segment, \ourmethod is bounding box-free, and can perform holistic image analysis with segmentation, detection, and recognition all at once. 
We conducted a large-scale evaluation on 102,855 test image-mask-label triples across 9 modalities. 
\ourmethod attained new state-of-the-art results, substantially outperforming prior best methods such as MedSAM and SAM, even when they were equipped with oracle bounding box as input. The improvement is even larger when the objects have irregular shapes or when an image contains a large number of objects. 
We also validated the accuracy and scalability of \ourmethod on previous unseen real-world data from Providence Health System. 
Collectively, \ourmethod offers an accurate, scalable, and robust biomedical image analysis tool that can be broadly applied to various modalities and applications, paving the path for image-based biomedical discovery.

The image analysis field has witnessed rapid development in the past decade.
Since its inception in 2015, the U-Net architecture has revolutionized the field of automatic pixel-wise prediction through supervised training \cite{ronneberger2015u, cciccek20163d}. This groundbreaking work laid the foundation for a diverse array of network structures, ranging from advanced convolution-network designs to vision-transformer models \cite{milletari2016v, li2018h,zhou2019unet++, myronenko20183d, isensee2021nnu,lee20223d, lee2023scaling, lee2023deformux, chen2021transunet, xu2023levit,xie2021cotr, wang2021transbts, hatamizadeh2022unetr, hatamizadeh2022swin, zhou2021nnformer, cao2021swin}. Recent advances in image detection and recognition, such as developments in object detection frameworks like Faster R-CNN \cite{ren2016faster} and YOLOv4 \cite{bochkovskiy2020yolov4}, have significantly enhanced capabilities in identifying and localizing anatomical features with high precision. 
The introduction of SAM marked a significant milestone by demonstrating the model's ability to generalize segmentation to previously unseen classes, utilizing visual prompts such as points and bounding boxes as guides \cite{kirillov2023segment}. 

Despite the proliferation of advances in the general domain, research on adapting them for large-scale biomedical image analysis across a wide range of organ or tissue classes remains relatively sparse ~\cite{Litjens_2017}. 
MedSAM is a notable exception by adapting SAM to the medical realm through continued training on a large number of biomedical segmentation datasets, establishing the state of art in biomedical image analysis.
However, like SAM, MedSAM focuses on segmentation alone, thus ignoring valuable semantic information from related tasks of detection and recognition. Consequently, both SAM and MedSAM require users to provide labor-intensive input such as the tight bounding box for each object to segment, which is hard to scale and very challenging for objects with irregular shapes ~\cite{ma2024segment}. 

We propose \ourmethod to overcome these challenges. By joint learning across segmentation, detection, and recognition in the unifying framework of image parsing, and by using GPT-4 to harmonize noisy object descriptions, \ourmethod was able to acquire novel capabilities such as identifying and segmenting objects of interest using text prompt alone, as well as recognizing all objects in an image by leveraging the segmentation ontology.
This represents an important step toward scaling holistic image analysis in biomedicine and real-world clinical applications.

A particularly exciting area for biomedical image analysis is the application in cellular images such as H\&E staining and Multiplexed ImmunoFluorescence (MxIF) imaging. This could help elucidate the size, shape, texture, and spatial relationships of individual cells, with potential ramifications in emerging applications such as modeling tumor microenvironments for precision immunotherapy \cite{stringer2021cellpose, greenwald2022whole, ma2023towards}. 
The standard approaches focus on instance segmentation by assigning unique identifiers to individual cells to facilitate downstream analysis \cite{girshick2015fast,he2017mask, schmidt2018cell}. Hover-net represents a significant advancement in addressing the limitations of semantic breadth and cell categorization within segmentation tasks, by incorporating cell classification into the segmentation process \cite{graham2019hover}. 
However, traditional methods typically rely on bounding box detection, and struggle with diverse cell morphologies and irregular shapes. 
Recent efforts aim to overcome these challenges by adopting more refined representations and accommodating the multi-resolution nature of biological imaging \cite{yang2020circlenet, nguyen2022circlesnake, ilyas2022tsfd}. Cell-ViT is a marquee example that leverages SAM's encoder backbone to improve hierarchical representation, particularly for nucleus segmentation \cite{horst2023cellvit}. 
\ourmethod can contribute to this long line of exciting research work by enabling cell segmentation and identification in one fell swoop and enhancing generalizability through joint training on a diverse range of image modalities and cell types.

While \ourmethod has demonstrated promising potential for unifying biomedical image analysis, growth areas abound. First, although \ourmethod has demonstrated high accuracy (e.g., Dice scores) in identifying relevant pixels in an image for a given object type, by default it does not differentiate individual object instances and requires post processing to separate the instance masks, which is important in some applications such as cell counting. Second, while \ourmethod can already perform image analysis from text prompt alone, it currently does not support interactive dialogue with users in a conversational style like GPT-4. To address this, we plan to develop a conversational system that can better tailor to complex user needs. Finally, \ourmethod currently treats non-2D modalities such as CT and MRI by reducing them to 2D slices, thus failing to utilize the spatial and temporal information in the original modalities. In future work, we need to extend \ourmethod beyond 2D image slices to facilitate 3D segmentation, detection, and recognition.

%% file: nbt_sec/methods.tex
\newpage 

\section*{Methods}
\subsection*{Details of \ourdata}

We created the first large-scale biomedical image parsing dataset \ourdata, where each image is associated with a collection of objects. Each object is annotated with the segmentation mask and a canonical semantic label specifying the object type from a biomedical object ontology. Additionally, each semantic label comes with a set of synonymous textual descriptions for model training.
\ourdata was created by synthesizing 45 publicly available biomedical segmentation datasets across 9 imaging modalities, comprising 1.1 million images, 3.4 million image-mask-label triples, and 6.8 million image-mask-description triples (\textbf{Fig. \ref{fig1:overview}b}). 
To ensure the quality of \ourdata, we imposed stringent inclusion criteria: each image had to be manually or semi-manually segmented at the pixel level, and a name was available for each segmented object from the dataset description. For 3D imaging modalities such as CT and MRI, we pre-processed each volume into in-plane 2D slices to be consistent with other modalities. 

For model training and evaluation, we randomly split each original dataset into 80\% training and 20\% testing. Slices from each 3D volume always appear in the same split to prevent information leakage. 

To harmonize natural-language variations in noisy object descriptions, we use GPT-4 to create a three-layer biomedical object ontology (\textbf{Fig. \ref{fig1:overview}a}). The base layer comprises three broad semantic categories: {\tt organ}, {\tt abnormality}, {\tt histology}. The next layer comprises 15 meta-object types (e.g., {\tt heart} in {\tt organ} and {\tt tumor} in {\tt abnormality}). The most fine-grained layer comprises 82 object types, such as {\tt left heart ventricle} and {\tt ehhancing tumor}.
Specifically, we first used GPT-4 to generate a preliminary hierarchical structure for biomedical image analysis and propose candidate names for individual object types, drawing from a wide range of tasks and textual descriptions across the source datasets. We then manually reviewed these candidates and mapped them to standardized OHDSI vocabularies using Athena \cite{ohdsi_athena_vocab}. 
We introduce {\tt other} as a catch-all category. For future expansion, we expect that the first two layers are relatively stable, while our framework can easily incorporate new object types in the fine-grained layers, as well as additional datasets with segmentation and object labels.


To enhance the robustness of \ourmethod in handling diverse text prompts, we also used GPT-4 to generate synonymous textual descriptions for each semantic label, following other recent efforts in using GPT-4 for synthetic data generation ~\cite{gu2023biomedjourney, li2024llava}. 
Specifically, we adopt a templatic normalization for each dataset, by formulating the unifying image analysis task as identifying ``[OBJECT TYPE] in [ANATOMIC SITE] [MODALITY]'', such as ``enhancing tumor in brain MRI'' (\textbf{Supplementary Figure} \ref{fig:data-modality-chord}).
We then introduced linguistic diversity into these descriptions by using GPT-4 to generate variations in professional language (\textbf{Supplementary Figure} \ref{fig:prompt}), as well as introducing synonymous variations for each component (\textbf{Supplementary Figure} \ref{fig:example}). 
In each training epoch, we randomly sampled a description for each image-mask pair, ensuring \ourmethod to understand diverse text prompts.

\subsection*{Details of \ourmethod}

Existing image analysis methods often focus on segmentation alone. They typically expect spatial input prompts such as bounding box or scribble for the object to segment, and focus on learning spatial embedding such as bounding box coordinates~\cite{kirillov2023segment,ma2024segment, wong2023scribbleprompt}.
In contrast, \ourmethod follows SEEM~\cite{zou2024segment} and focuses on learning text prompt. 
Specifically, \ourmethod adopts a modular design, comprising an image encoder, a text encoder, a mask decoder, and a meta-object classifier. See \textbf{Fig. \ref{fig1:overview}c}. The image and text encoders were initialized using Focal~\cite{yang2022focal} and PubMedBERT ~\cite{gu2021domain}, respectively. 

The input to \ourmethod is an image and a text prompt, which are passed along to the image and text encoders, respectively. The text prompt specifies the object type for segmentation and detection.
The mask decoder outputs a segmentation mask that has the same size as the original image, with a probability between 0 and 1 for each pixel, indicating how likely the pixel belongs to the designated object in the text prompt. 
The meta-object classifier includes input from the image and outputs the meta-object type to facilitate joint training of image encoder with object semantics. 

\subsection*{Implementation of competing methods}
We compared \ourmethod to state-of-the-art segmentation models, SAM \cite{kirillov2023segment} and MedSAM \cite{ma2024segment}. We recognize the importance of precise bounding boxes as model input, so we evaluated competing methods in two settings:
(i) employing gold-standard bounding boxes, and (ii) utilizing bounding boxes predicted by the state-of-the-art object detection model Grounding DINO \cite{liu2023grounding} to provide bounding box prompts.
For the first setting, we follow \cite{ma2024segment} by deriving bounding boxes from gold-standard masks, ensuring each box tightly encompassed the mask with a uniform margin of 10 pixels.
In the second setting, we adhered to the inference pipeline of Grounding DINO where when presented with multiple bounding box predictions, we selected the one with the highest confidence score. This text-to-box-to-segmentation scheme follows the idea in \cite{ren2024grounded}.
To maintain uniformity across comparisons, all input images were resized to $1024 \times 1024$ pixels.
We use the same test split of \ourdata for evaluation across competing methods, and performance was quantified using the median Dice score on each task. We recognize that the train-test splits are different across the original evaluations of the competing methods, and the \ourdata test split could contain examples that were used to train other models.
We note that the implementations for MedSAM, SAM, and Grounding DINO were used as-is for inference purposes without any fine-tuning. As for the task-specific nnU-Net models \cite{isensee2021nnu} and the DeepLabV3+ models \cite{chen2018encoder}, due to the unavailability of numerous task-finetuned models and the lack of explicit training details in existing literature, we relied on performance metrics reported in the MedSAM study \cite{ma2024segment}.


\subsection*{Detecting invalid textual description}

\ourmethod by design can input any image and text prompt. However, a text prompt may be invalid, specifying an object that doesn't exist in the given image~\cite{li2024llava,lee2023ai}. For example, the request to identify and segment ``left heart ventricle'' in a dermoscopy image should be rejected by the model as invalid. 
It is critical to detect and reject invalid text prompt to preempt hallucinations~\cite{achiam2023gpt}. 

In principle, the mask decoder should output low pixel probabilities for invalid text prompt. However, given the sheer number of pixels, some might get a relatively high output probability simply by chance, thus leading to erroneous object detection and segmentation results. To address this problem, we observe that while individual pixels might get noisily high probabilities, collectively their distribution would be rather different compared to pixels in valid objects. Consequently, we can estimate the distribution of its pixel probabilities from training data, and then estimate how likely the pixel probabilities in a test image are drawn from the same distribution. 

Specifically, after \ourmethod was trained, for each object type, we computed the average object pixel probability for each training image containing objects of the given type, and fit a Beta distribution for all these probabilities.
At test time, for a given image, we computed the average object pixel probability for the predicted object segments of the given object type, and compute the $p$-value using one-sample Kolmogorov-Smirnov (K-S) test \cite{massey1951kolmogorov}.
Smaller $p$-value indicates that the predicted object segments are unlikely to be correct.
To increase the robustness, in addition to pixel probability, we also consider the RGB values. In particular, for each color channel (R, G, B), we similarly fit a Beta distribution from the average value for valid objects in training, and compute the corresponding $p$-value for the predicted object segments in a test image.
Overall, we treat these four tests as independent and use their product as the summary $p$-value. 

In this way, we can obtain a summary $p$-value for any given pair of text prompt and image. 
To identify a summary $p$-value threshold for separating valid inputs from invalid ones, we created an invalid dataset by sampling invalid object types for each image. 
We plot the distribution for both valid text prompts (for a given image) and invalid ones (\textbf{Fig.~\ref{fig2:quant_eval}f}).
For comparison against Grounding DINO, we use its confidence score given a text prompt and an image for invalid input detection.


\subsection*{Attention map conditioned on the textual description}
To visualize the shape of each segmentation object type, e.g. ``hepatic vessel in CT'', we collected the predicted pixel probabilities for each object type and aggregated probabilities from all images. The pixel-level probability is derived from the top layer attention on the pixel. The attention map, reflecting the shape for a target $t$, is obtained in a four-step approach. First, we collected all \ourmethod-predicted pixel attention for target $t$ as $\rho_1, \cdots, \rho_n \in [0,1]^{H\times W}$ across $n$ examples in the test set. Second, we initialized shape distribution for target $t$ as $\mathcal{M}^t_1 = \rho_1$. Third, for iteration $i=1, \cdots, n-1$, we computed 2-D cross-correlation between $\rho_{i+1}$ and $\mathcal{M}^t_i$, and shifted $\rho_{i+1}$ to be aligned with $\mathcal{M}^t_i$ at highest cross-correlation, and updated the ensemble distribution $M_{i+1}^t = M_i^t + \tilde{\rho}_{i+1}$, where $\tilde{\rho}_{i+1}$ denotes the shifted attention matrix. Finally, the attention map for target $t$ is normalized as $M_n^t / n$. For 3D segmentation targets such as CT and MRI, we first aggregated the predictions within one volume without shifting and then aligned the volume-aggregated masks using the above method.

\subsection*{Details of experiments on irregular-shaped object detection}
Medical image segmentation models like MedSAM require a bounding box as input. When the shape of the target is ``irregular'', it is hard for the bounding box to precisely define the region of interest. To quantify the ``regularity'' of a target mask $M$, we define the following three metrics: \textbf{Box Ratio} measures the degree to which the target mask is similar to its tight bounding box: $ BoxRatio(M) = \frac{|M|}{|Box(M)|}$, where $Box(M)$ is the tight bounding box around mask $M$, and $|\cdot|$ denotes the area measured in number of pixels. \textbf{Convex Ratio} measures how convex the target mask is and is defined as $ConvexRatio(M) = \frac{|M|}{|ConvexHull(M)|}$,  where $ConvexHull(M)$ is the convex hull of mask $M$. \textbf{Inverse rotational inertia} (IRI) measures how spread out the area of the target mask is. To begin with, the rotational inertia of $M$ relative to its centroid $c_M$ is $ RI(M) = \sum_{x \in M} \|x - c_M\|_2^2$, where $x$ is the coordinate of each pixel in the mask, and $c_M$ is the coordinate of the centroid. To standardize the metric to be independent of the total mask area, we take the inverse of the rotational inertia and scale by the value of a round-shaped mask with the same area, representing the lowest rotational inertia achievable by any mask with the same area: $IRI(M) = \frac{|M|^2}{2\pi \cdot RI(M)}$.
Under this definition, any mask has $0 < IRI \leq 1$, with any round-shaped mask having IRI equal to 1.

\subsection*{Details of experiments on object recognition}
We built a hierarchical structure putting all supported targets under one modality at one anatomic site. Given any image, e.g. abdominal CT, we traverse all the available targets $t = 1, \cdots, m$ under the branch that are exclusive to each other, and prompt the \ourmethod model sequentially to get $m$ prediction of mask probabilities $\rho^1, \cdots, \rho^m$. It is possible that the predicted masks can overlap with each other. The challenges then are how to select the right set of targets in the specific image and how to determine the right mask regions for the selected targets to avoid overlapping. We used a two-stage approach for object recognition, including a target selection stage and a mask aggregation stage. In the target selection stage, we first calculate the original mask area for each target $t$ as $A^t$. Then, we iterate through the pixels. For each pixel $(i,j)$, we rank the targets that have pixel probability $\rho^t_{ij} > 0.5$. The target assigned to pixel $(i,j)$ is $T_{ij} = \mbox{argmax}\, \rho^{t'}_{ij}.$ After this round of pixel assigning, the final area for each target $t$ is $\tilde{A}^t = \sum_{i,j} \mathbf{1}_{T_{ij}=t}$. The targets with final area $\tilde{A}^t > \lambda A^t$ are the selected targets, with $\lambda$ being the user-specified threshold. In the mask aggregation stage, we discard all unselected target masks completely, and then iterate through the pixels again. For each pixel, the most probable target $t$ with $\rho^t_{ij} > 0.5$ is assigned. The pixels with predicted probabilities $\rho^t_{ij} \leq 0.5$ for all selected targets are left blank. 

For the baseline method using Grounding DINO with SAM and MedSAM, we first prompted Grounding DINO with the set of targets to retrieve a collection of bounding boxes with confidence scores. Then we implemented non-maximum suppression \cite{canny1986computational, viola2001rapid, girshick2014rich} to select a subset of identified targets in the scene, minimizing the overlapping between the targets. To get the segmentation masks for these identified targets, we further prompted SAM and MedSAM with the bounding boxes to retrieve the corresponding predictions.


%% file: data_code_arxiv.tex
\section*{Data availability}
We will provide access to \ourdata or scripts to reproduce \ourdata from the original datasets, upon publication of this manuscript.

\section*{Code availability}
\ourmethod will be made fully available upon publication, including the model weights and relevant source code for pre-training, fine-tuning, and inference. We will also provide detailed methods and implementation steps to facilitate independent replication.

%% file: nbt_sec/figures.tex
\newpage
\renewcommand*{\figurename}{\textbf{Figure}}
\begin{figure}[!ht]
    \centering
    \includegraphics[width=.90\textwidth]{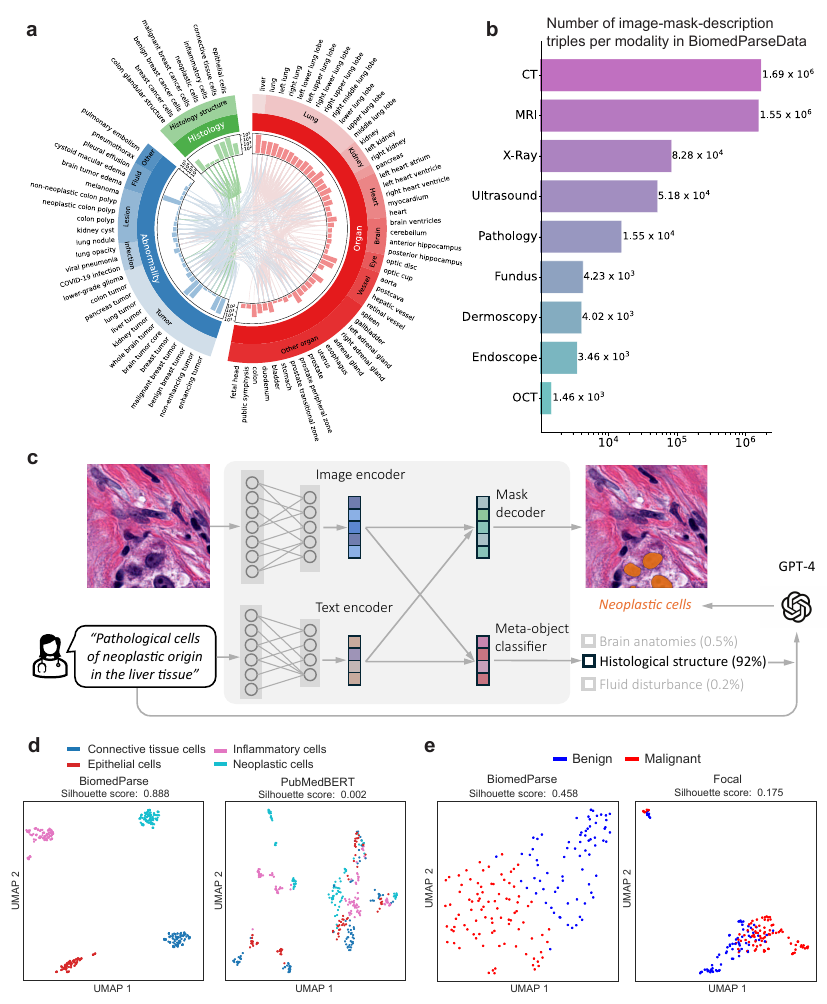}
    \caption{\textbf{Overview of \ourmethod and \ourdata.} \textbf{a, } The GPT-4 constructed ontology showing a hierarchy of object types that are used to unify semantic concepts across datasets. 
    Bar plots showing the number of images containing that object type. Lines between bars showing the object type similarity in the text embedding space. \textbf{b,} Bar plot showing the number of image-mask-description triples for each modality in \ourdata. \textbf{c,} Flowchat of \ourmethod. \ourmethod takes an image and a text prompt as input and then outputs the segmentation masks for the objects specified in the prompt. Image-specific manual interaction such as bounding box or clicks is not required in our framework. To facilitate semantic learning for the image encoder, \ourmethod also incorporates a learning objective to classify the meta-object type. 
    For evaluation, GPT-4 is used to resolve text prompt into object types using the object ontology, which also uses the meta-object type output from \ourmethod to narrow down candidate semantic labels.     
    \textbf{d,} UMAP plots contrasting the text-embeddings for different cell types derived from \ourmethod text encoder (left) and PubMedBERT (right). \textbf{e,} UMAP plots contrasting the image embeddings for different cell types derived from \ourmethod image encoder (left) and Focal (right). }
    \label{fig1:overview}
\end{figure}


\begin{figure}[tbp]
    \centering
    \includegraphics[width=0.9\textwidth]{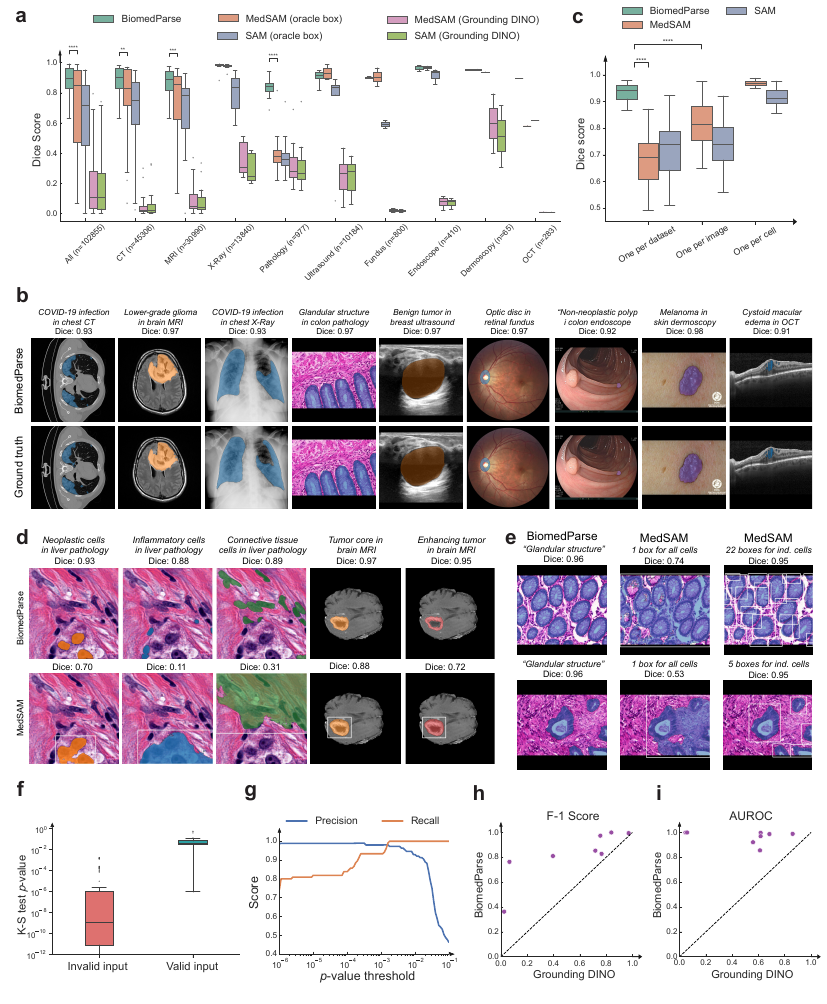}
    \caption{\textbf{Comparison on large-scale biomedical image segmentation datasets.} ~\textbf{a}, Bar plot comparing the Dice score between our method and competing methods on 102,855 test instances (image-mask-label triples) across 9 modalities. 
    MedSAM and SAM require bounding box as input. We consider two settings: oracle bounding box (minimum bounding box covering the gold mask); bounding boxes generated from the text prompt by Grounding DINO, a state-of-the-art text-based grounding model. 
    $n$ denotes the number of test instances in the corresponding modality. $*$ indicates the significance level at which {\ourmethod} outperforms the best-competing method, with Wilcoxon test $p$-value$<1\times10^{-2}$ for **, $p$-value$<1\times10^{-3}$ for ***,  $p$-value$<1\times10^{-4}$ for ****. \textbf{b}, Nine examples comparing the segmentation results by \ourmethod and the ground truth, using just the text prompt at the top. \textbf{c}, Bar plot comparing the Dice score between our method and competing methods on a cell segmentation test set with 42 images. \ourmethod requires only a single user operation (the text prompt ``Glandular structure in colon pathology''). By contrast, to get competitive results, MedSAM/SAM require 430 operations (one bounding box per an individual cell).}
    \label{fig2:quant_eval}
\end{figure}
\clearpage
\captionsetup{type=figure} 
\caption*{\textbf{d}, Five examples contrasting the segmentation results by \ourmethod and MedSAM, along with text prompts used by \ourmethod and bounding boxes used by MedSAM. ~\textbf{e}, Comparison between \ourmethod and MedSAM on a benign tumor image (top) and a malignant tumor image (bottom). The improvement of \ourmethod over MedSAM is even more pronounced on abnormal cells with irregular shapes. ~\textbf{f}, Bar plot comparing the K-S test $p$-values between valid text prompt and invalid text prompt. \ourmethod learns to reject invalid text prompts describing object types not present in the image (small $p$-value). ~\textbf{g}, Plot showing the precision and recall of our method on detecting invalid text prompts across different K-S test $p$-value cutoff. ~\textbf{h,i}, Scatter plots comparing the AUROC (\textbf{h}) and F-1 (\textbf{i}) between \ourmethod and Grounding DINO on detecting invalid descriptions.}

\begin{figure}[tbp]
    \centering
    \includegraphics[width=0.95\textwidth]{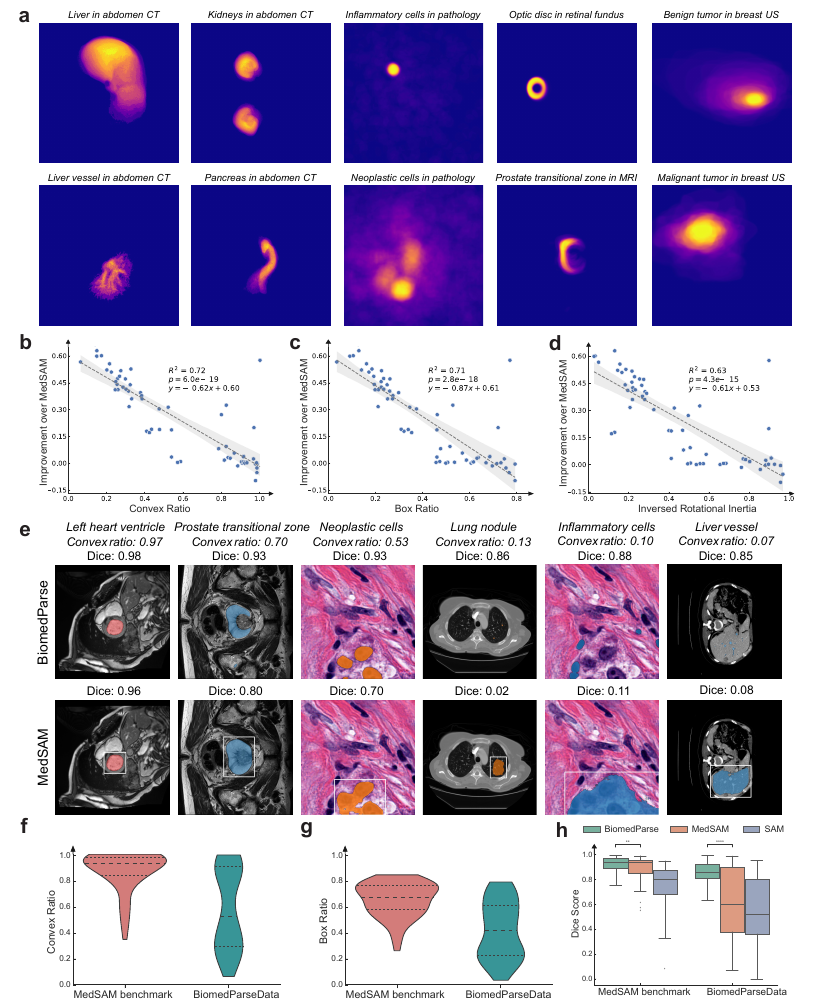}
    \caption{\textbf{Evaluation on detecting irregular-shaped objects.} \textbf{a,} Attention maps of text prompts for irregular-shaped objects, suggesting that \ourmethod learns rather faithful representation of their typical shapes. \textbf{b-d, } Scatter plots comparing the improvement in Dice score for \ourmethod over MedSAM with shape regularity in terms of convex ratio (\textbf{b}), box ratio  (\textbf{c}), and inversed rotational inertia (\textbf{d}). Smaller number in x-axis means higher irregularity in average. Each dot is an object type. \textbf{e,} Six examples contrasting \ourmethod and MedSAM on detecting irregular-shaped objects. Plots are ordered from the least irregular one (left) to the most irregular one (right). \textbf{f,g} Comparison between \ourdata and the benchmark dataset used by MedSAM in terms of convex ratio (\textbf{f}) and box ratio (\textbf{g}). \ourdata is a more faithful representation of real-world challenges in terms of irregular-shaped objects. \textbf{h,} Bar plots comparing \ourmethod and competing approaches on \ourdata and the benchmark dataset used by MedSAM. \ourmethod has a larger improvement on \ourdata, which contains more diverse images and more irregular-shaped objects.}
    \label{fig3:irreg_eval}
\end{figure}

\begin{figure}[tbp]
    \centering
    \includegraphics[width=0.95\textwidth]{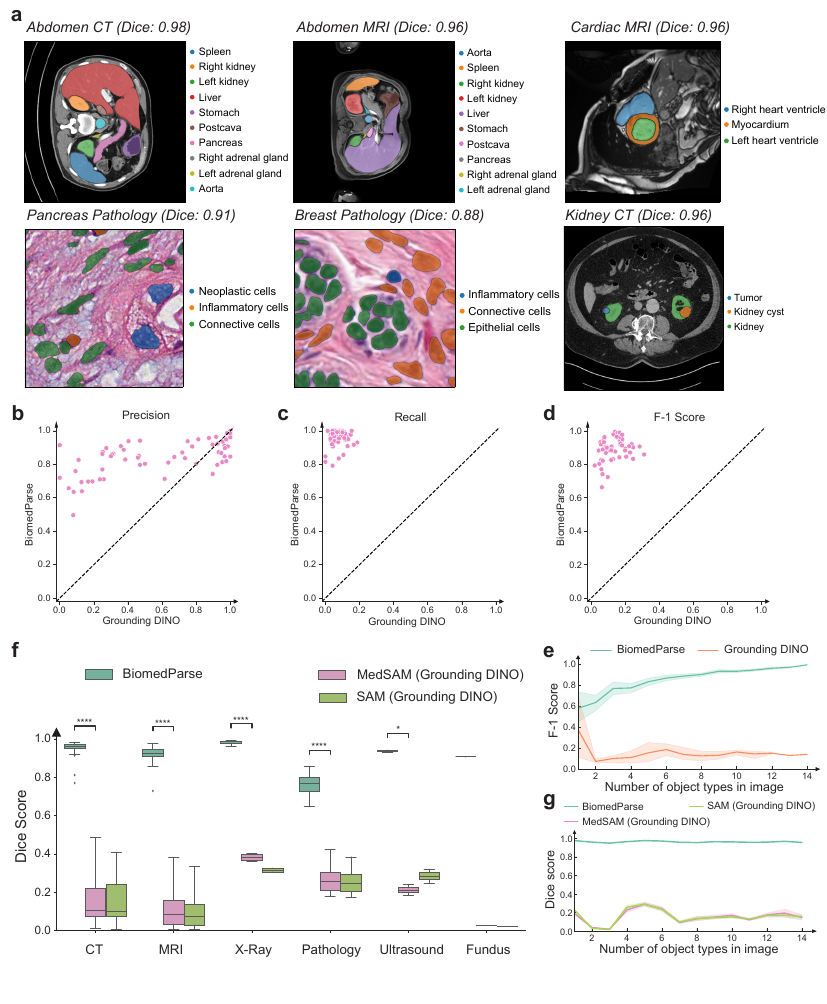}
    \caption{\textbf{Evaluation on object recognition.} \textbf{a,} Six examples showing the results of object recognition by our method. Object recognition identifies and segments all objects in an image without requiring any user-provided input prompt. \textbf{b-d, } Scatter plots comparing the F1 (\textbf{b}), Precision (\textbf{c}), and Recall (\textbf{d}) scores between \ourmethod and Grounding DINO on identifying objects presented in the image. \textbf{e, } Comparison between \ourmethod and Grounding DINO on object identification in terms of median F-1 score across different numbers of objects in the image. \textbf{f, } Bar plot comparing \ourmethod and MedSAM/SAM (using bounding boxes generated by Grounding DINO) on end-to-end object recognition (including segmentation) in relation to various modalities. \textbf{g, } Comparison between \ourmethod and MedSAM/SAM (using bounding boxes generated by Grounding DINO) on end-to-end object recognition (including segmentation) in relation to numbers of distinct objects in the image. }
    \label{fig4:pano_eval}
\end{figure}

\newpage

\begin{figure}
    \centering
    \includegraphics[width=\textwidth, trim={0 2.5cm 0 0}, clip]{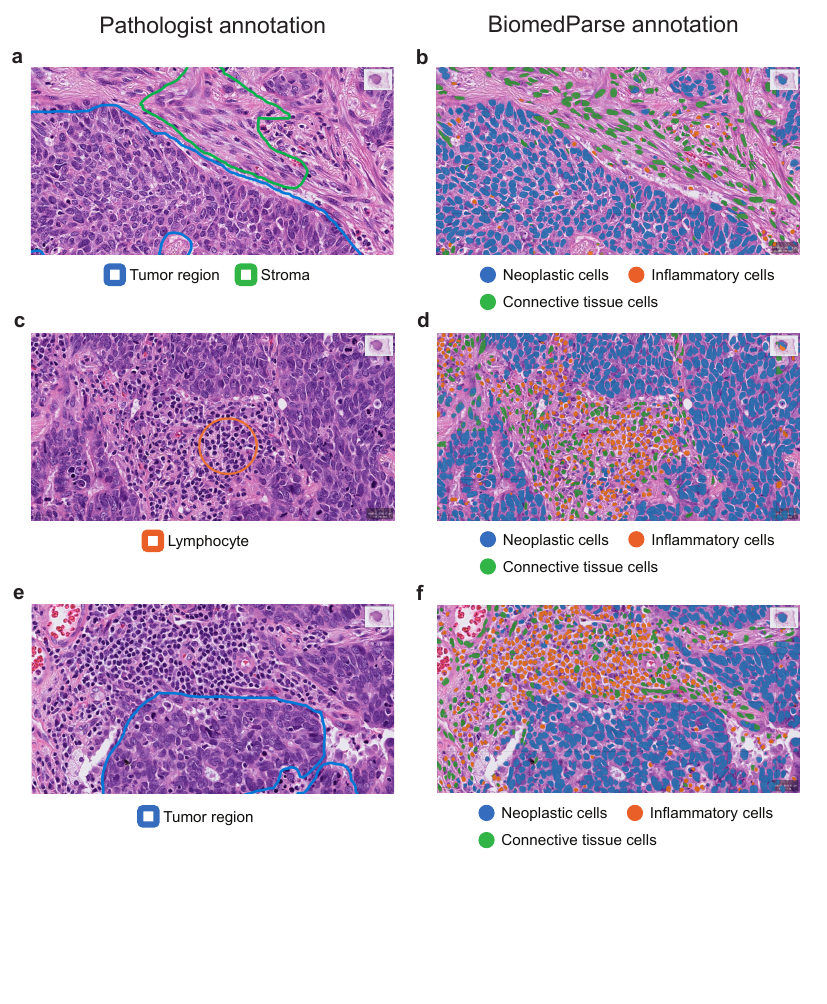}
    \caption{\textbf{Evaluation of \ourmethod on real-world cell segmentation examples.} \textbf{a-f}, De-identified pathology images from Providence Health System are used to compare the pathologist annotations (\textbf{a,c,e}) and the annotations from \ourmethod (\textbf{b,d,f}). We show the exact pathologist outputs, including object names (e.g., lymphocyte, stroma) and object locations, as well as the exact outputs by \ourmethod. \ourmethod does not need any user-provided inputs and can identify and segment cells of any types included in the ontology.}
    \label{fig5:prov_ex}
\end{figure}


%% file: nbt_sec/supplementary.tex
\newpage

\setcounter{figure}{0}
\renewcommand*{\figurename}{\textbf{Supplementary Figure}}

\begin{figure}[tbp]
    \centering
    \includegraphics[width=0.9\textwidth, trim={0 2cm 0 0}, clip]{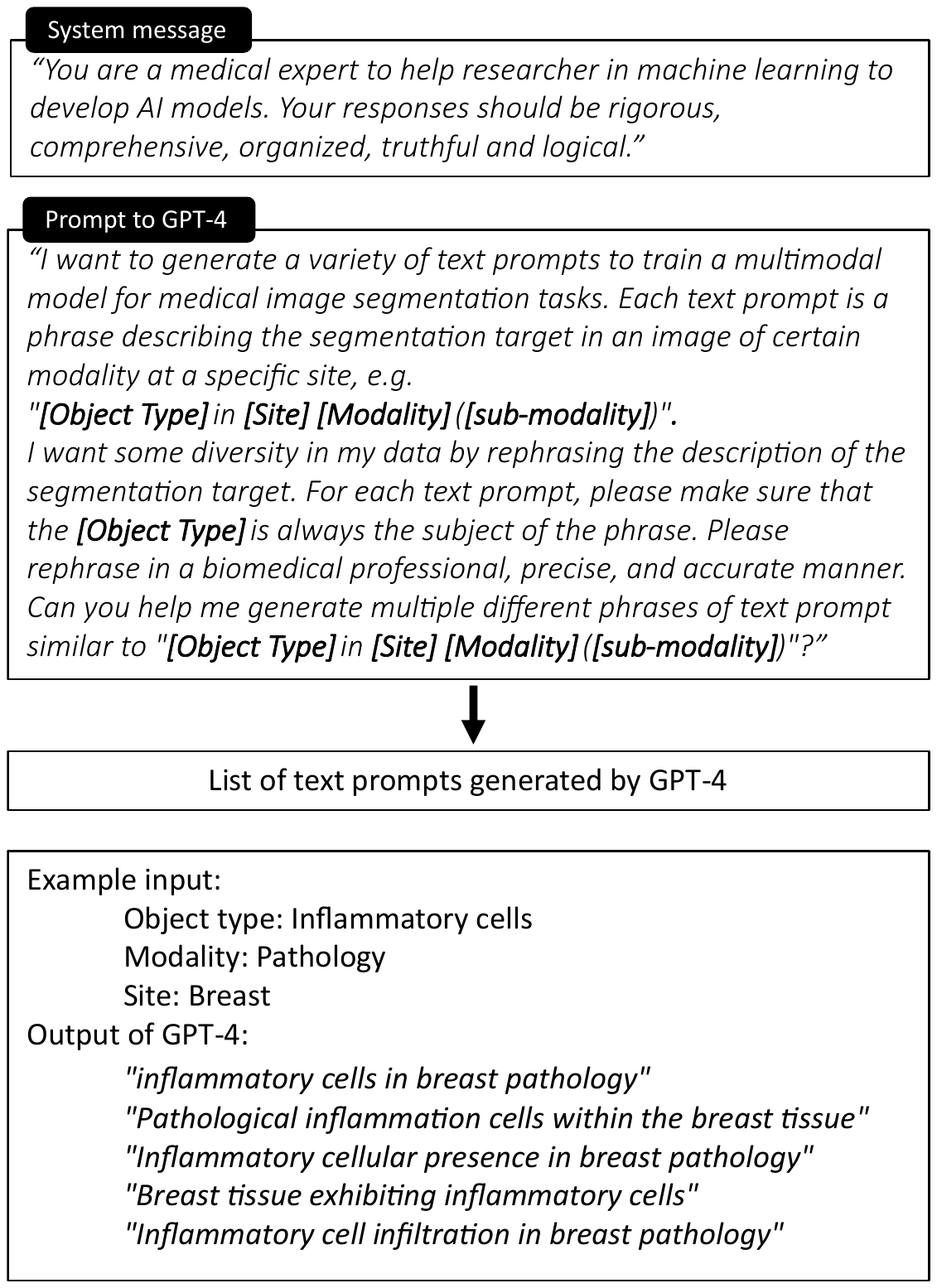}
    \caption{GPT-4 prompt that is used to generate diverse descriptions for a given image according to its object type, image modality, and anatomic site.}
    \label{fig:prompt}
\end{figure}

\newpage

\begin{figure}[tbp]
    \centering
    \includegraphics[width=.95\textwidth]{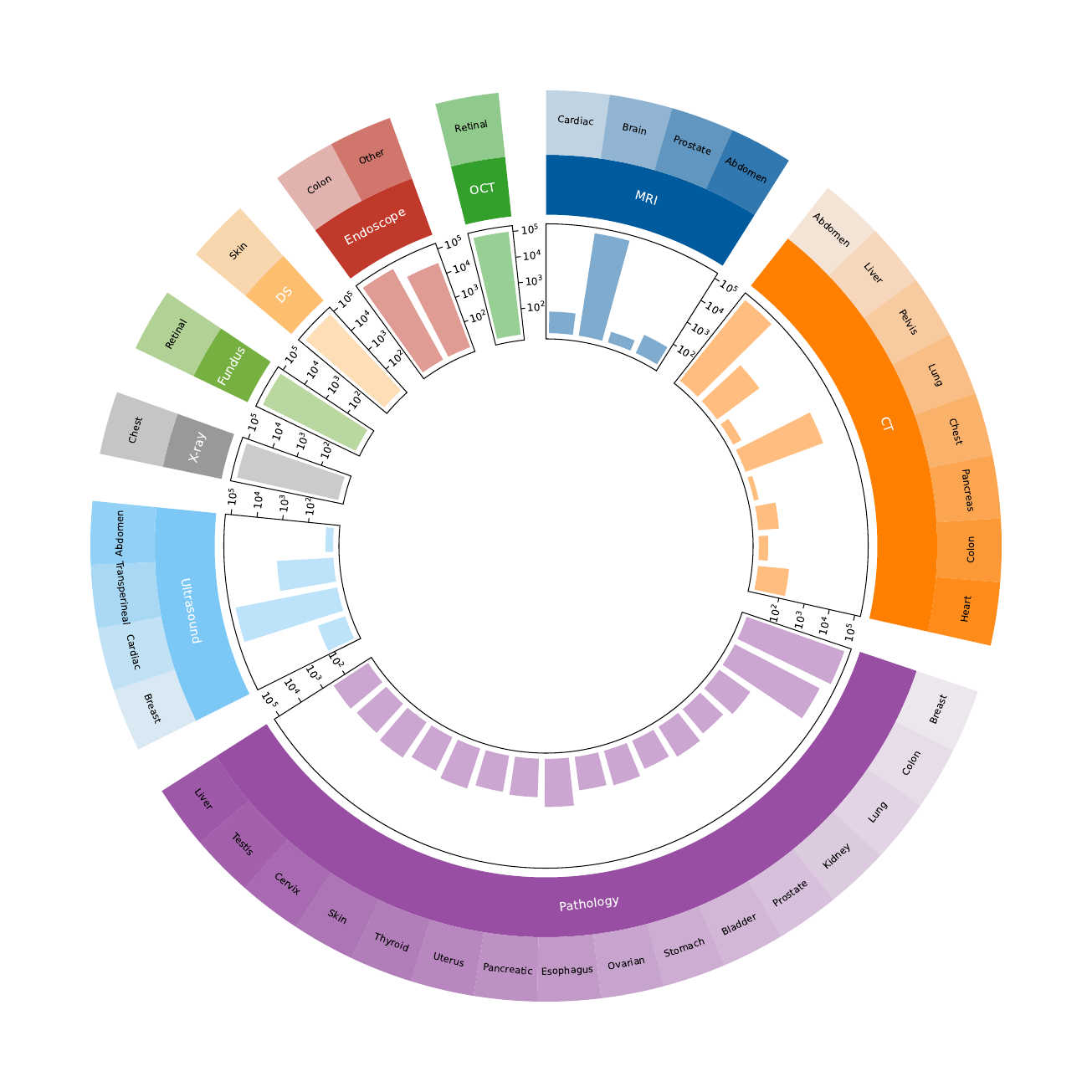}
    \caption{Number of images in each of the 25 anatomic sites from 9 modalities. One anatomic site could present in multiple modalities.}
    \label{fig:data-modality-chord}
\end{figure}

\newpage

\begin{figure}[!ht]
    \centering
    \includegraphics[width=0.95\textwidth, trim={0 7cm 0 0}, clip]{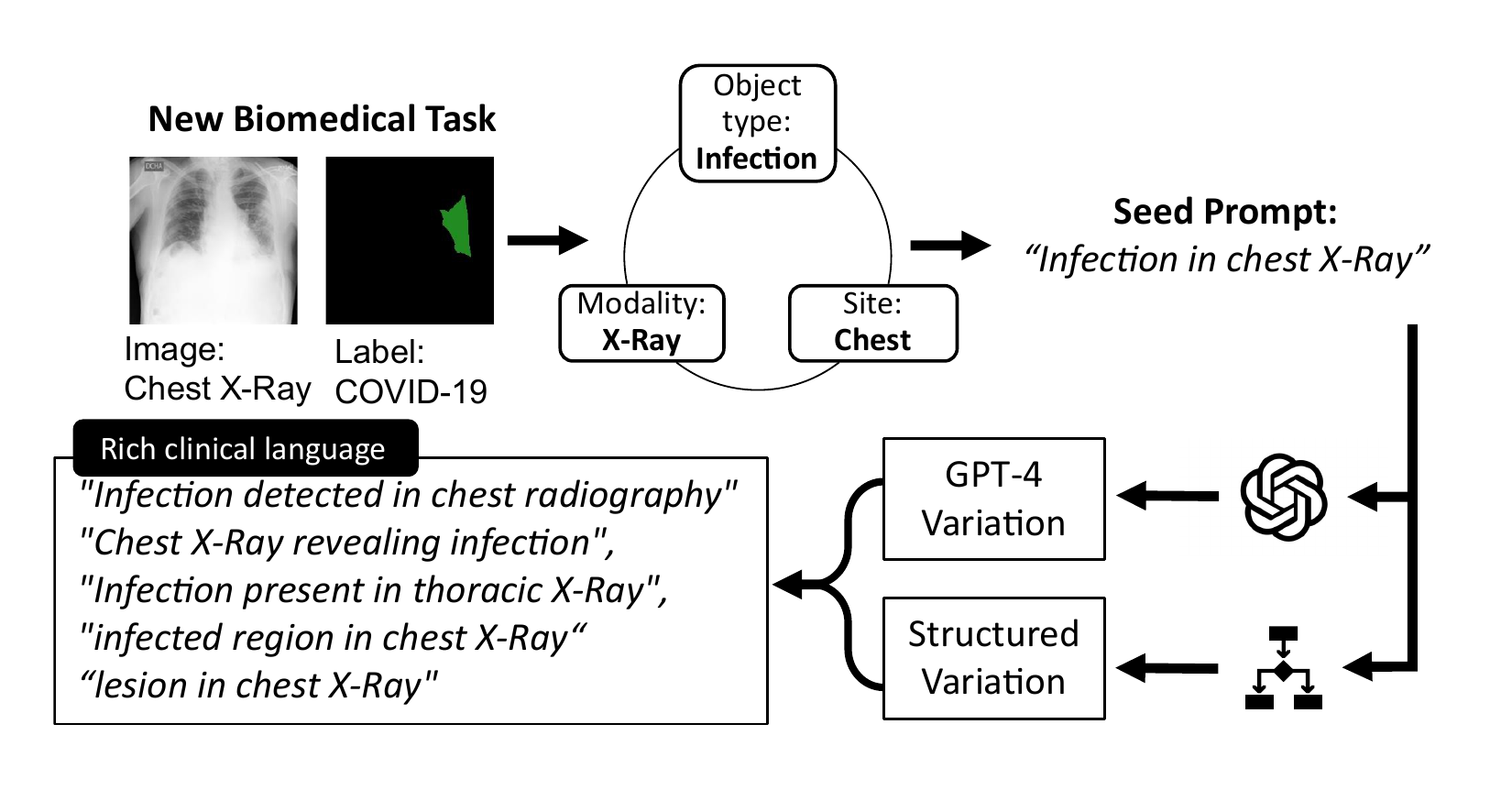}
    \caption{Generating textual description for each object in each image. Object type, modality, and site are extracted from the metadata or the data description. We utilized both GPT-4 and structured biomedical concepts to generate rich variations of clinical language, increasing the robustness of \ourmethod to user-provided text.}
    \label{fig:example}
\end{figure}

\newpage

\begin{figure}[tbp]
    \centering
    \includegraphics[width=.95\textwidth]{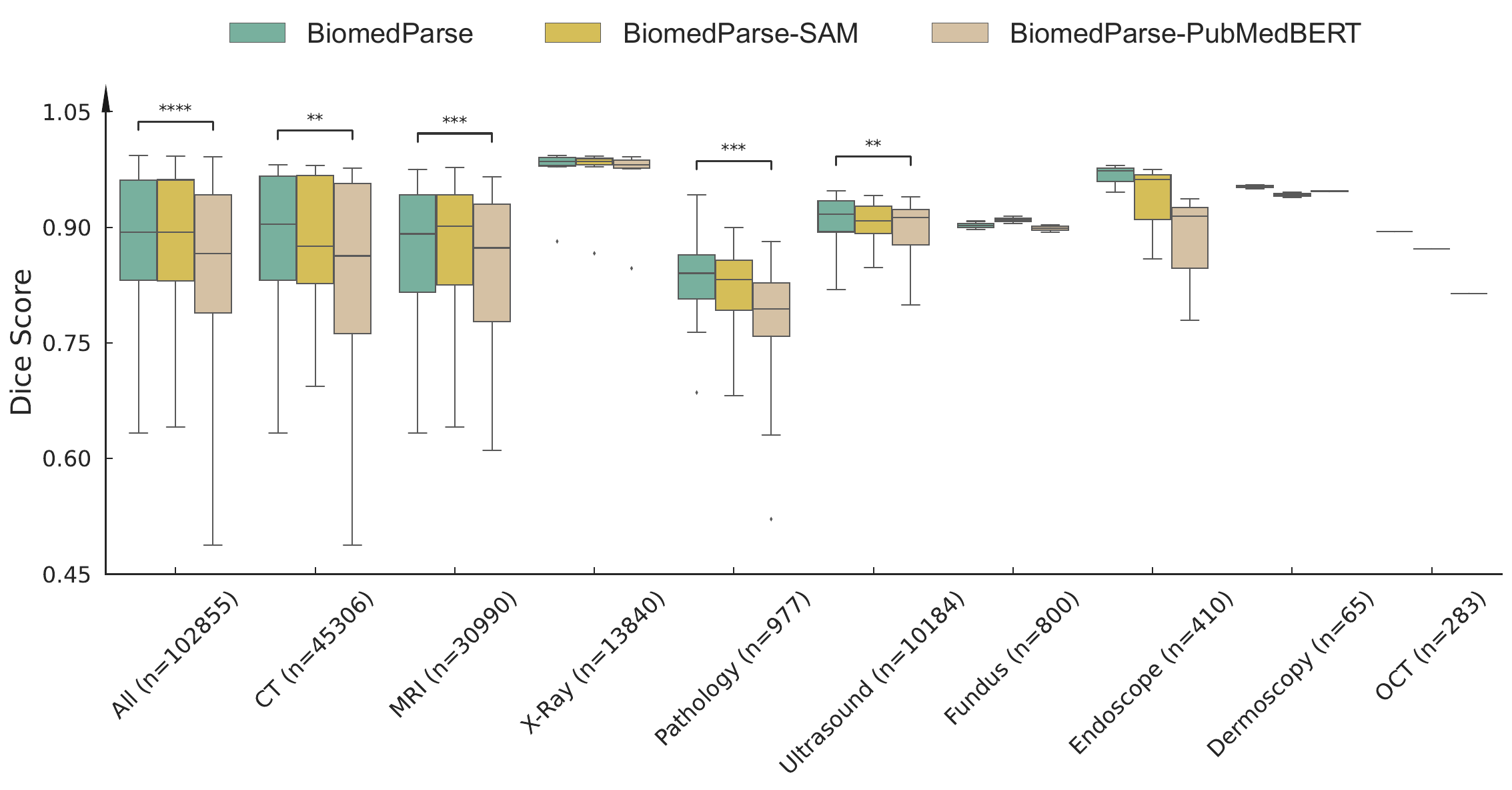}
    \caption{Ablation studies comparing the performance of \ourmethod and two variants. \ourmethod-SAM stands for using SAM to initialize the image encoder. \ourmethod-PubmedBERT stands for using the frozen PubmedBERT \cite{gu2021domain} as the text encoder. $n$ denotes the number of images in the corresponding modality. $*$ indicates the significance level at which {\ourmethod} outperforms the best-competing method, with Wilcoxon test $p$-value$<1\times10^{-2}$ for **, $p$-value$<1\times10^{-3}$ for ***,  $p$-value$<1\times10^{-4}$ for ****.}
    \label{fig:ablation}
\end{figure}

\newpage

\begin{figure}[tbp]
    \centering
    \includegraphics[width=.95\textwidth]{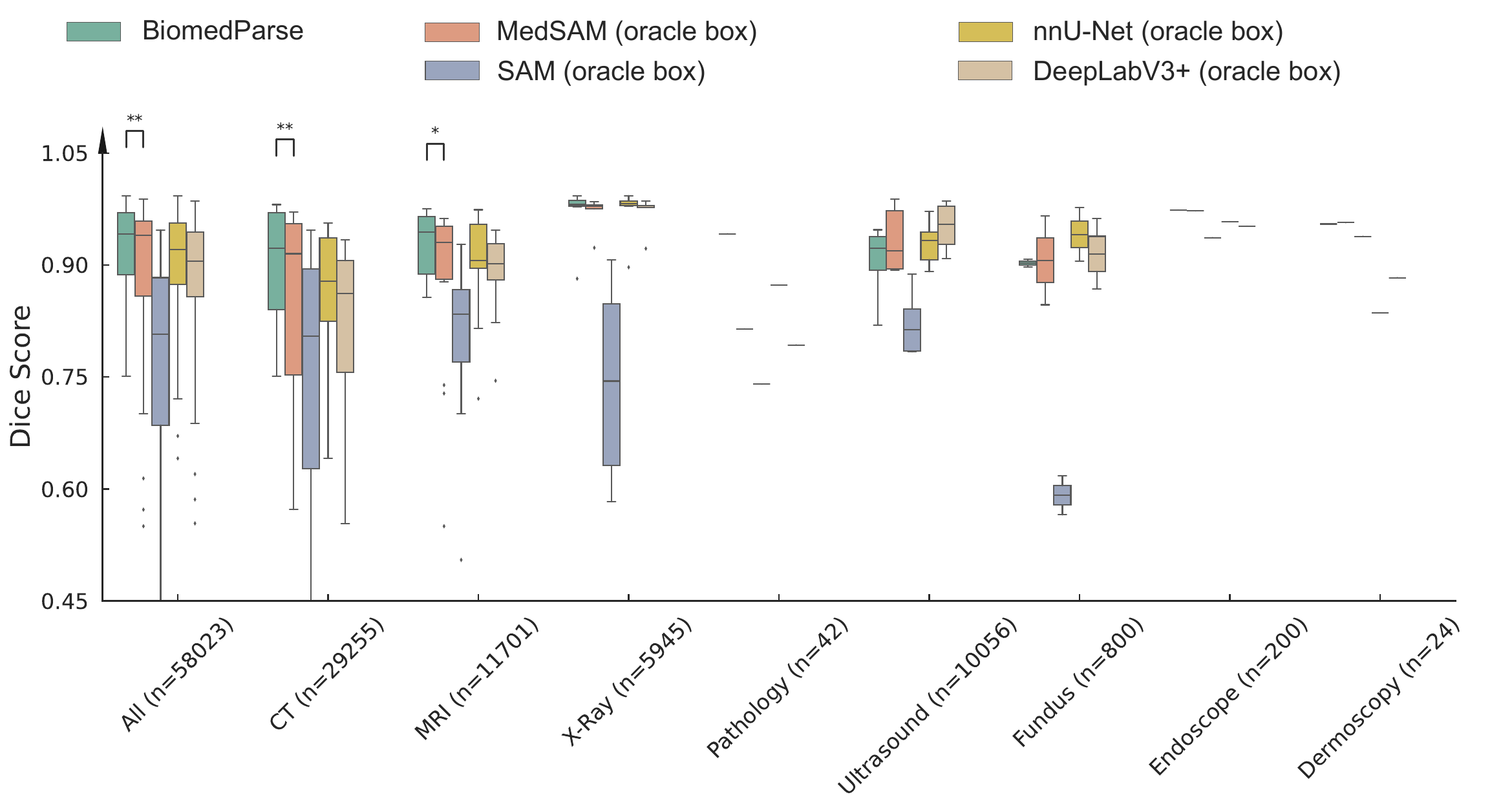}
    \caption{Comparison between \ourmethod and competing methods on the MedSAM benchmark. We evaluated MedSAM and SAM using the ground truth bounding box for the segmentation. For nnU-Net and DeepLabV3+, we reported the evaluation reported by MedSAM \cite{ma2024segment}. Results are shown by imaging modality, with statistical significance comparison between \ourmethod and best-competing method MedSAM. $*$ indicates the significance level at which {\ourmethod} outperforms the best-competing method, with Wilcoxon test $p$-value$<5\times10^{-2}$ for *, $p$-value$<1\times10^{-3}$ for **.}
    \label{fig:overlap_eval_mod}
\end{figure}

\newpage



\begin{figure}[tbp]
    \centering
    \includegraphics[width=.32\textwidth]{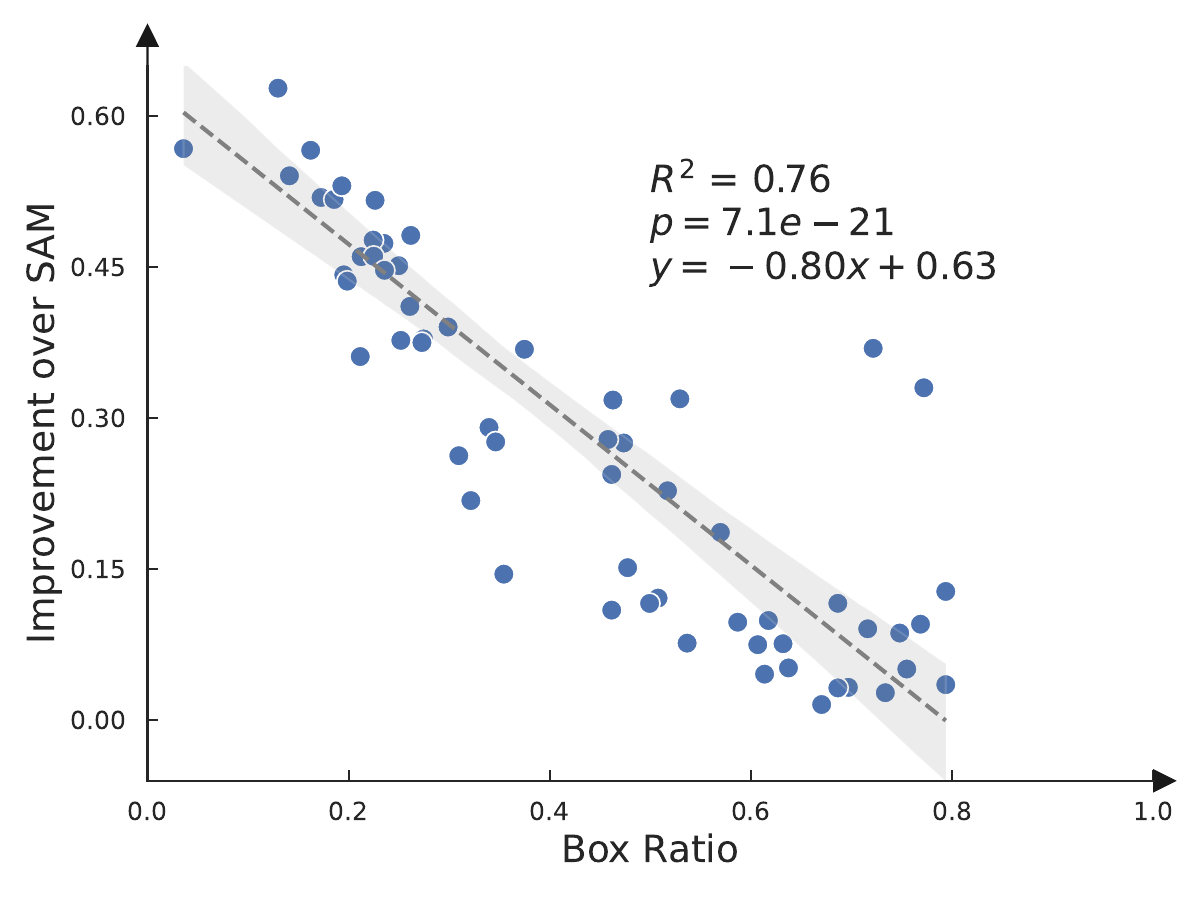} 
    \includegraphics[width=.32\textwidth]{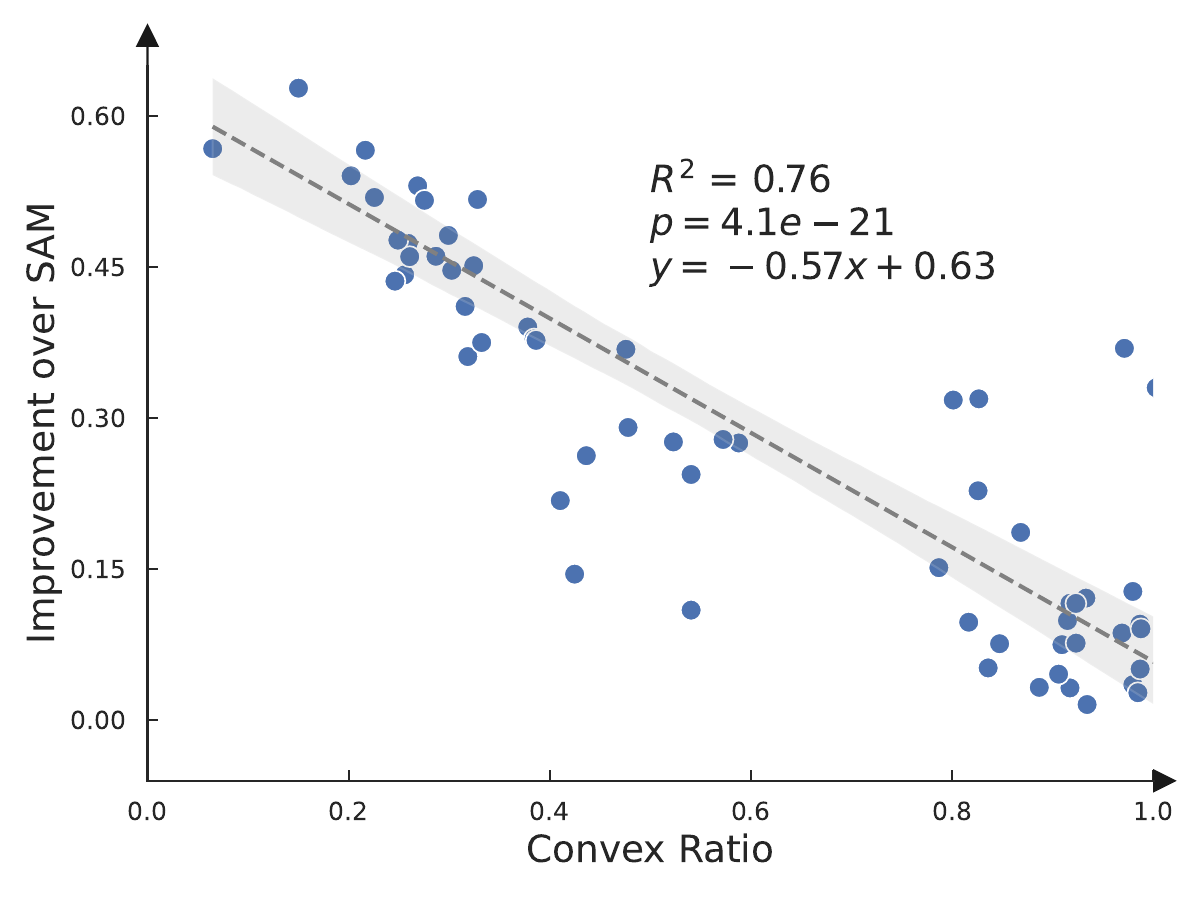} 
    \includegraphics[width=.32\textwidth]{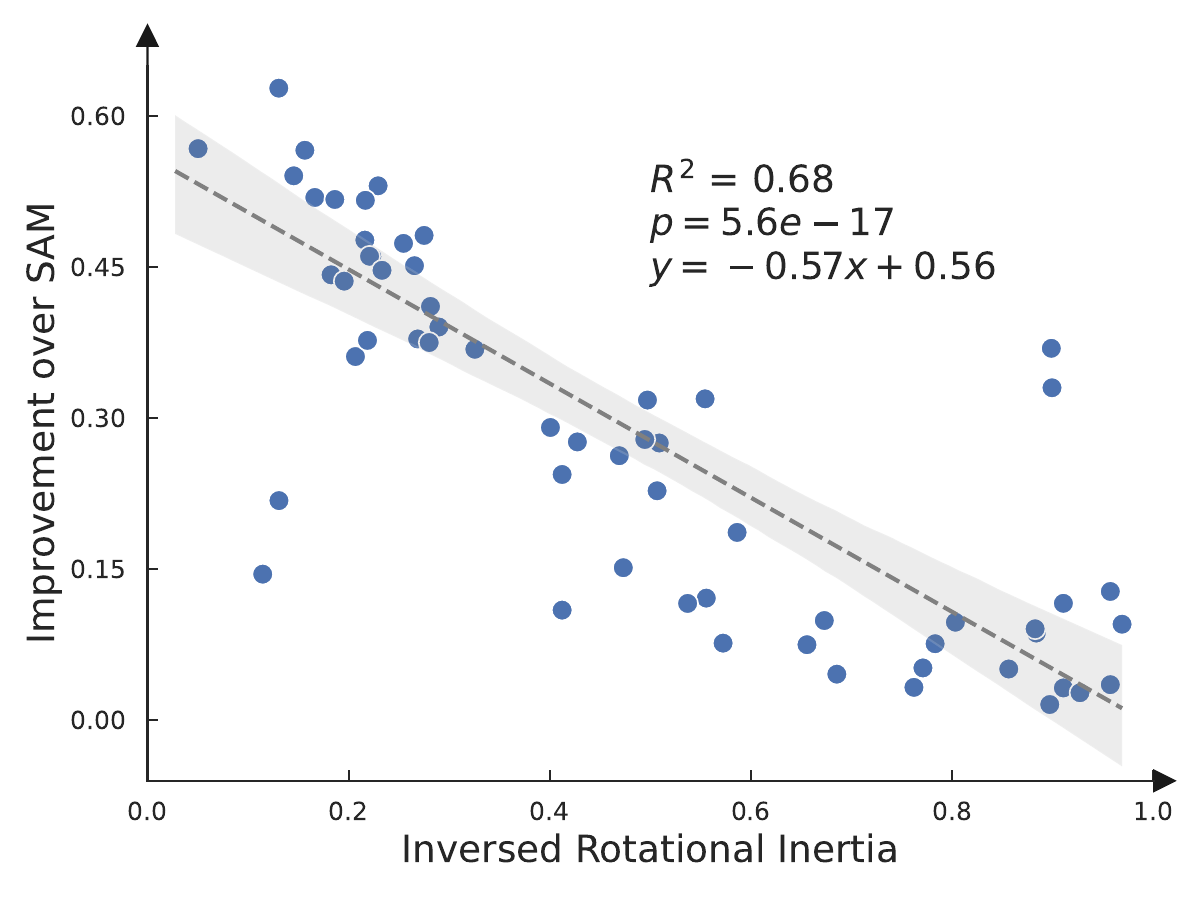}
    \caption{Scatter plots comparing the improvement of \ourmethod over SAM with shape irregularity in terms of box ratio (left), convex ratio  (middle), and inversed rotational inertia (right). Each dot represents the median statistics over one object type in our segmentation ontology.}
    \label{fig:irregular_medsam}
\end{figure}

\newpage

\begin{figure}[tbp]
    \centering
    \includegraphics[width=.5\textwidth]{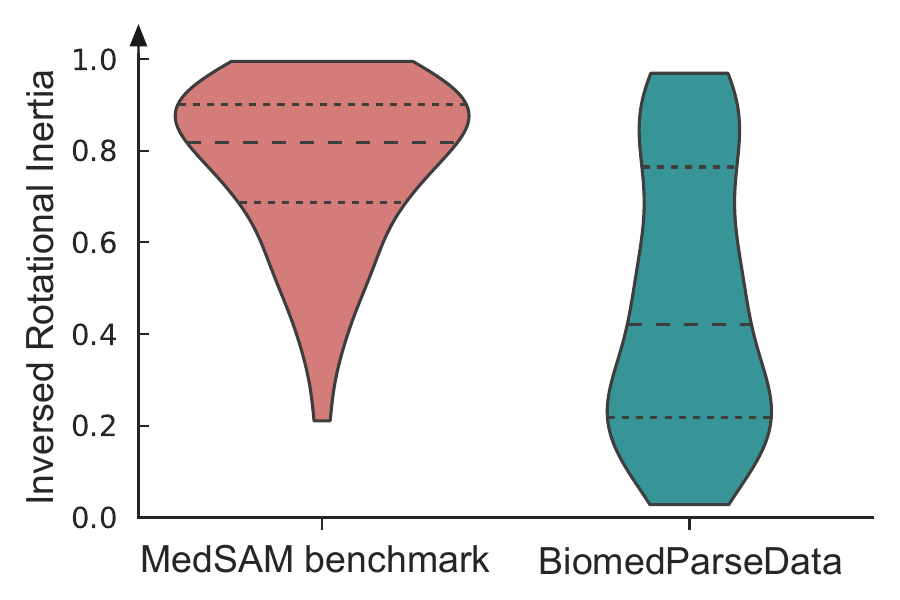}
    \caption{ Violin plot comparing the inversed rotational inertia between MedSAM benchmark data and \ourdata. A higher inversed rotational inertia indicates less irregularity. }
    \label{fig:IRI_violin}
\end{figure}

\newpage

\begin{figure}[tbp]
    \centering
    \includegraphics[width=.49\textwidth]{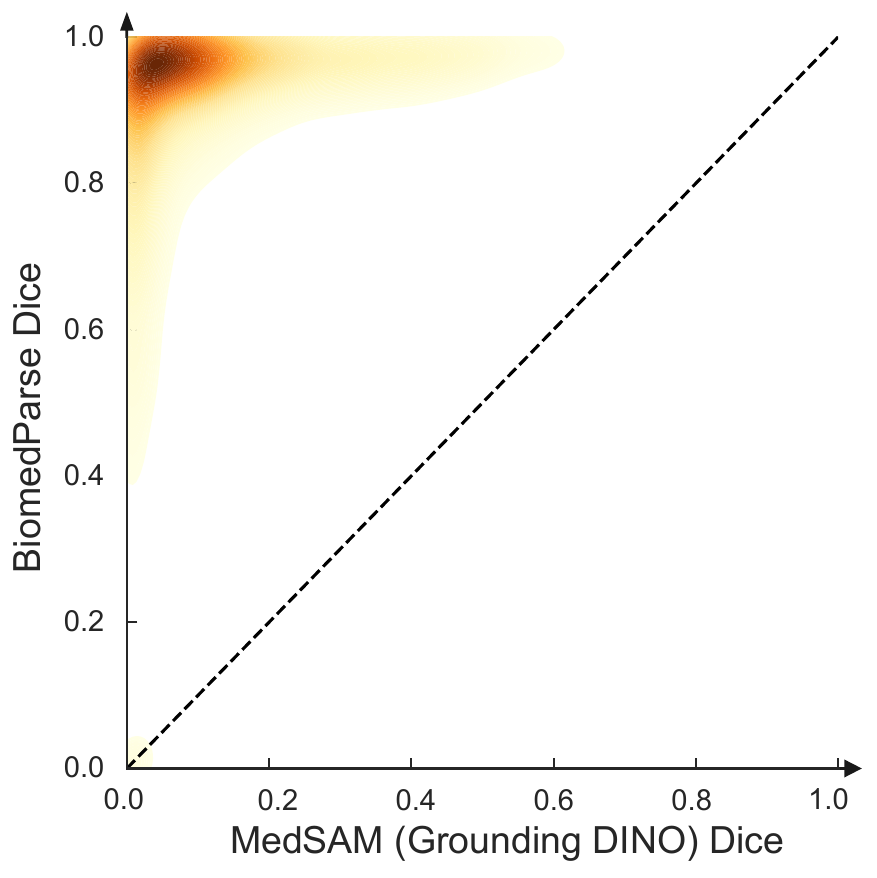}
    \includegraphics[width=.49\textwidth]{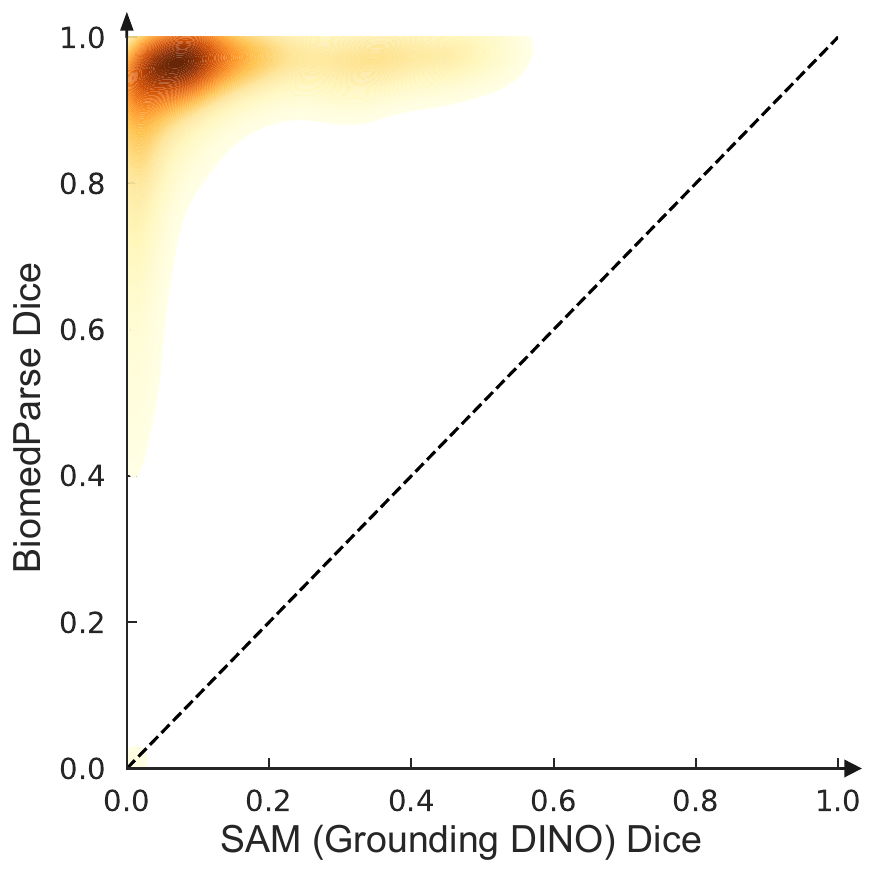}
    \caption{\textbf{Evaluation on object recognition.}\textbf{a,b,} Density plots comparing the performance on object recognition between \ourmethod and MedSAM (Grounding DINO) \textbf{a}, and between \ourmethod and SAM (Grounding DINO) \textbf{b}.}
    \label{fig:panoptic_scatter}
\end{figure}

\newpage